\documentclass{article}

\usepackage{spconf}
\usepackage{epsfig}
\usepackage{subcaption}
\usepackage{calc}
\usepackage{cite}
\usepackage{amsmath,amssymb,amsfonts}
\usepackage{algorithmic}
\usepackage{graphicx}
\usepackage{textcomp}
\usepackage{xcolor}
\usepackage{tabularx}
\usepackage{here}
\usepackage[inline]{enumitem}
\usepackage{hyperref}
\usepackage{relsize}
\usepackage{makecell}
\usepackage{array}

\newif\ifcamera
\cameratrue           

\begin{document}

\ifcamera
\title{Leveraging NeRF-Rendered Images for 3D Gaussian Splatting}

\name{Mizuki Morikawa, Yuta Shimizu, Chunyu Li, Yusuke Monno, Masatoshi Okutomi}
\address{Institute of Science Tokyo, Tokyo, Japan}

\else
\title{Leveraging NeRF-Rendered Images for 3D Gaussian Splatting}
\name{Anonymous Authors}
\address{}
\fi

\maketitle

\begin{abstract}
Neural radiance field (NeRF) and 3D Gaussian splatting (3DGS) are two mainstream approaches for novel view synthesis. They often show complementary performance, i.e., 3DGS demonstrating faster rendering speed and NeRF demonstrating higher rendering quality. Motivated by this, we propose leveraging NeRF-rendered images for 3DGS. Specifically, we target street scenes and utilize a pre-trained street-specific NeRF method to produce training images for a target 3DGS method. In our 3DGS training, NeRF-rendered images are used to remove transient objects in street-level input views and to generate bird's-eye views as additional views, inheriting the higher-quality rendering of NeRF into 3DGS. We further incorporate a diffusion-based image enhancement to improve the image quality of the additional views. Experimental results on one synthetic and two real datasets demonstrate that our proposed method improves street-scene rendering while preserving the speed of 3DGS and the quality of NeRF.
\end{abstract}


\section{Introduction}
Novel view synthesis~(NVS) is the task of rendering images from arbitrary viewpoints by modeling an explicit or implicit 3D representation of the scene from observed multi-view images.
3D Gaussian splatting (3DGS)~\cite{gaussian_splatting} and neural radiance fields~(NeRF)~\cite{nerf} are currently two mainstream approaches for NVS. It is widely understood that 3DGS has a strong advantage in rendering speed and can achieve real-time NVS, whereas NeRF usually cannot. On the other hand, whether 3DGS or NeRF achieves higher rendering quality is inconclusive due to their differences in 3D representations and rendering mechanisms.

\begin{figure}[t!]
      \centering
\includegraphics[width=1.00\linewidth]{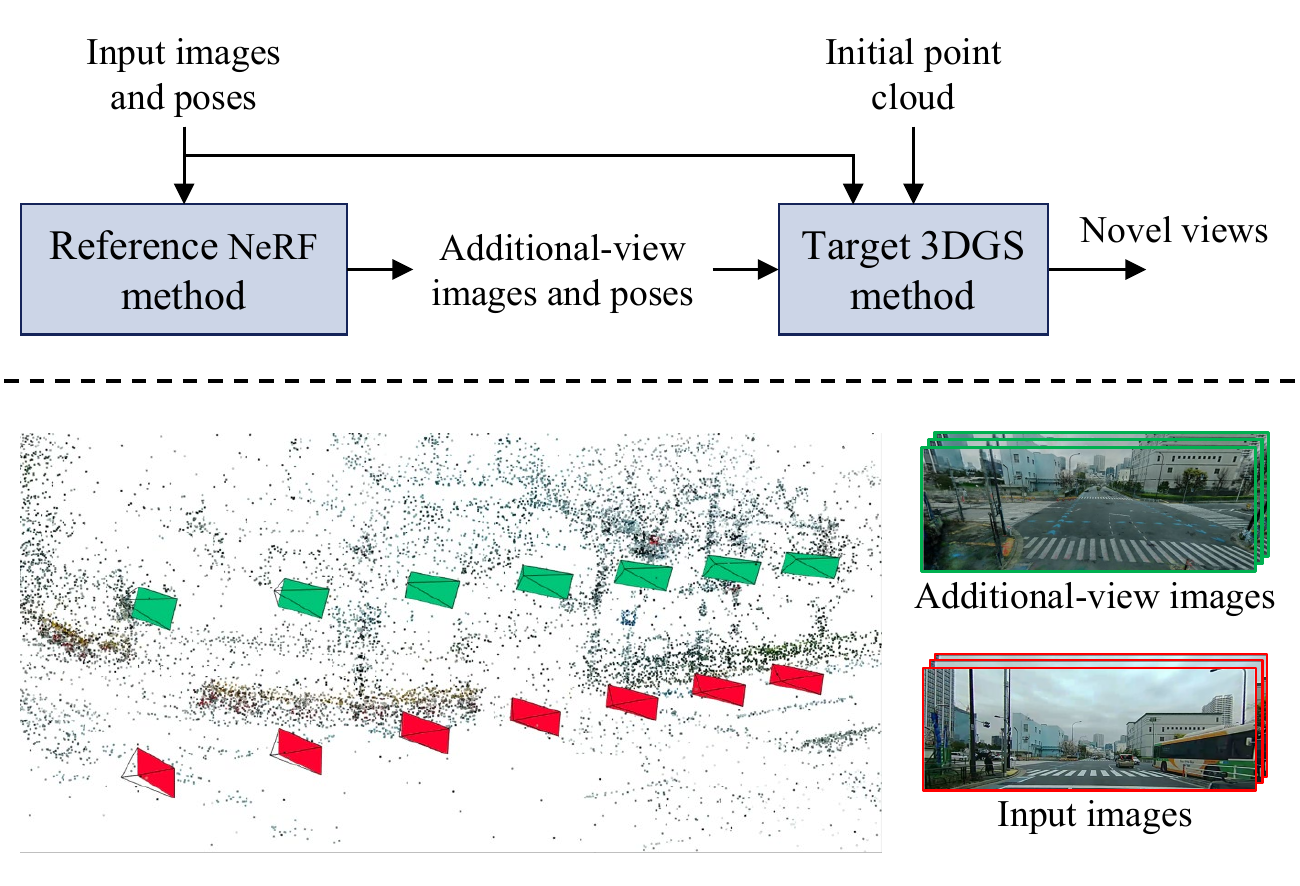}\\ \vspace{-3mm}
      \caption{The top part shows the general and conceptual flow of our proposed method, where we utilize NeRF-rendered images for additional views to train a 3DGS method. The bottom part shows the examples of generated additional views targeting the NVS of a street scene.}
      \vspace{-2mm}
\label{fig:overview_general}
\end{figure}

In this study, we target the situation where 3DGS and NeRF provide complementary performance, i.e., 3DGS achieves real-time rendering, and NeRF demonstrates higher rendering quality. In this situation, we propose leveraging NeRF-rendered images to train 3DGS. The top of Fig.~\ref{fig:overview_general} shows the general and conceptual flow of our proposed method. We first train a NeRF method, which is called reference NeRF, in a standard manner using input images and corresponding camera poses. Then, we render additional-view images from generated camera poses using the reference NeRF model. They are subsequently used for training a 3DGS method, which is called target 3DGS, as well as the original input images and an initial point cloud estimated by structure from motion~(SfM)~\cite{schoenberger2016sfm}. In this pipeline, the reference NeRF plays a role in augmenting training views for the target 3DGS, exploiting its higher rendering quality.

\begin{figure*}[t!]
      \centering
\includegraphics[width=1.00\linewidth]{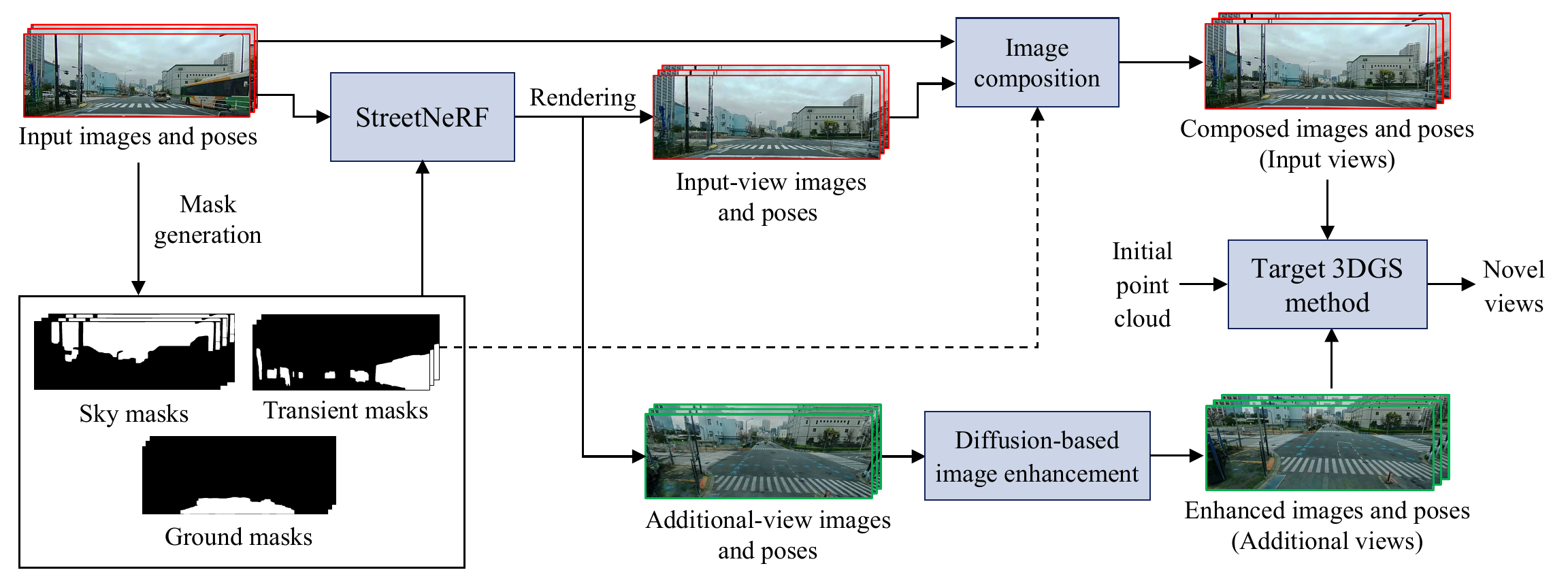}\\ \vspace{-2mm}
      \caption{The overview of our proposed method using StreetNeRF as the reference NeRF method, targeting NVS in street scenes. 
      }
\label{fig:overview}
\end{figure*}

Similar to this study, some recent studies addressed combining NeRF and 3DGS. Viewing direction Gaussian
splatting~(VDGS) introduces viewing direction encoding of NeRF into the opacity modeling of 3D Gaussians~\cite{vdgs}. NeRF-GS proposes a framework to jointly train NeRF and 3DGS to realize feature-level coupling between them~\cite{nerf-gs}. While these studies design a specific network model for each method, we propose a more general approach that exploits a pre-trained reference NeRF model to augment training images for 3DGS, allowing the inclusion of arbitrary 3DGS-based methods as the target 3DGS.

With the above feature of our method in mind, in this paper, we design a method specifically tailored for NVS in street scenes. For this purpose, we employ a recent segmentation-guided NeRF method for street scenes~\cite{streetnerf}~(we call it StreetNeRF in short) and generate bird's-eye views as additional views, as shown at the bottom of Fig.~\ref{fig:overview_general}, where red cameras are original input views from a car-mounted camera and green cameras are generated additional views. By adding the bird's-eye views rendered from StreetNeRF in high quality, our method improves the NVS quality of the target 3DGS while preserving its real-time performance. Furthermore, inspired by the recent study of~\cite{difix3d+}, we apply a diffusion-based image enhancement to the additional views as an option, which further enhances the NVS quality. In experiments, we apply our proposed method to three representative 3DGS methods, i.e., Splatfacto~\cite{gsplat} (3DGS implementation in Nerfstudio~\cite{nerfstudio}), Scaffold-GS~\cite{scaffoldgs}, and 2DGS~\cite{2dgs}, and evaluate them using one synthetic and two real datasets. The results demonstrate that our method consistently improves the NVS quality of the above target 3DGS methods, especially for novel views downward to the street, which are unobservable from street-level input views. 

\section{Proposed Method}


\subsection{Overview}

Figure~\ref{fig:overview} illustrates the overview of our proposed method using StreetNeRF~\cite{streetnerf} as the reference NeRF method, targeting NVS of street scenes. Firstly, following StreetNeRF, three types of binary masks are generated using Grounded SAM~\cite{grounded_sam} to identify the sky, the ground, and the transient objects in each input image. Then, StreetNeRF is trained using those masks, input images, and camera poses estimated by SfM~\cite{schoenberger2016sfm}. The trained NeRF model is then used to render images at the input views and newly defined bird's-eye additional views. For the input views, image composition is performed using the transient masks to generate transient-object-removed composed images. For the additional views, a diffusion-based image enhancement~\cite{difix3d+} is optionally applied to generate enhanced images. Finally, the composed images and the enhanced images are used as the inputs to train a target 3DGS method, enabling the generation of novel view images.


\subsection{Review of StreetNeRF}

StreetNeRF is a derivative of Zip-NeRF~\cite{zipnerf} specifically adapted to street scenes by introducing segmentation-guided region decomposition and region-specific processing to address challenging areas that typically lead to performance degradation in street scenes. Using the generated three types of binary masks mentioned before, specialized treatments are applied to each region, including infinite-depth color modeling for the sky, plane regularization for the ground, and disabling loss computation in transient-object regions. In addition, to mitigate artifacts caused by inconsistencies in illumination conditions across input images, color transformations for each image are trained for appearance embedding.

From the aspect of our proposed method, StreetNeRF provides the following benefits.
\begin{itemize}
    \vspace{-2mm}
    \item \textbf{Input-view images}: Since StreetNeRF excludes the transient-object regions from loss computation during its training, the rendered images for input views (i.e., training views) are not affected by transient objects. Because these images also contain information behind moving objects, they enable the target 3DGS method to be trained with richer visual information.
    \vspace{-2mm}    
    \item \textbf{Additional-view images}: Additional information not directly observed from the input views can be supplied to the target 3DGS method. By generating effective additional views for street scenes, a higher-quality rendering of StreetNeRF, according to the plane regularization and the sky color modeling, can be inherited into the target 3DGS method.
\end{itemize}
The detailed processes are explained below.


\subsection{Image composition for input views}

At the input views, replacing the real images entirely with the images rendered from the trained StreetNeRF model can suppress the influence of transient objects. However, this may reduce the reliability of the image content in static regions. To balance these factors, we compose the two types of images to construct better input-view images. For this purpose, we reuse the binary transient mask to replace only the regions corresponding to the transient objects in the real images with those rendered by the NeRF model. This allows us to utilize high-confidence images for training the target 3DGS method while eliminating the influence of the transient objects.

\begin{figure}[t!]
      \centering
\includegraphics[width=1.0\linewidth]{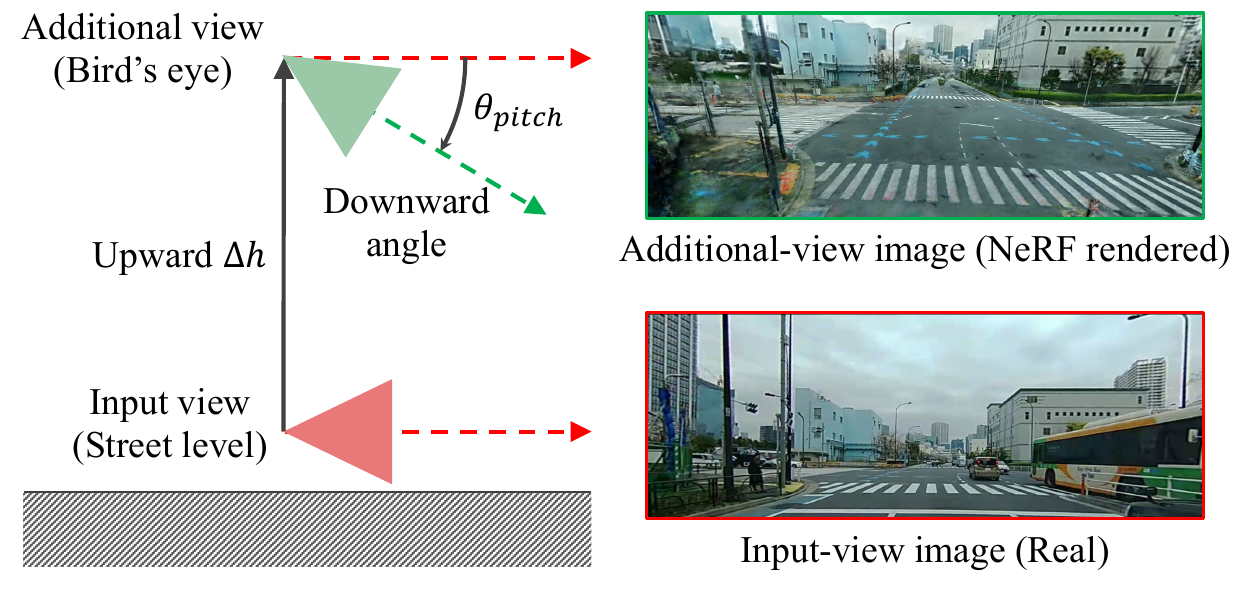}\\ \vspace{-2mm}
      \caption{Schematic illustration of additional view generation.}
\label{fig:added_pose}
\end{figure}


\subsection{Generation of additional views}

In street scenes, typical viewpoints that cannot be observed using an on-board camera, but are important for NVS applications such as street map viewing, are the viewpoints looking down at the street from the top. Thus, we generate bird’s-eye views as additional views, as shown in Fig.~\ref{fig:added_pose}. The parameters we set are the upward distance $\Delta h$ from the input views and the downward angle $\theta_{pitch}$ for the bird's-eye views.

As an optional process to further enhance the image quality, an image enhancement diffusion model provided in~\cite{difix3d+} is applied to the NeRF-rendered additional-view images. This process sharpens the regions that are blurry in the images.

\section{Experimental Results}


\subsection{Results on synthetic dataset}

\begin{figure}[t!]
      \centering
\includegraphics[width=1.0\linewidth]{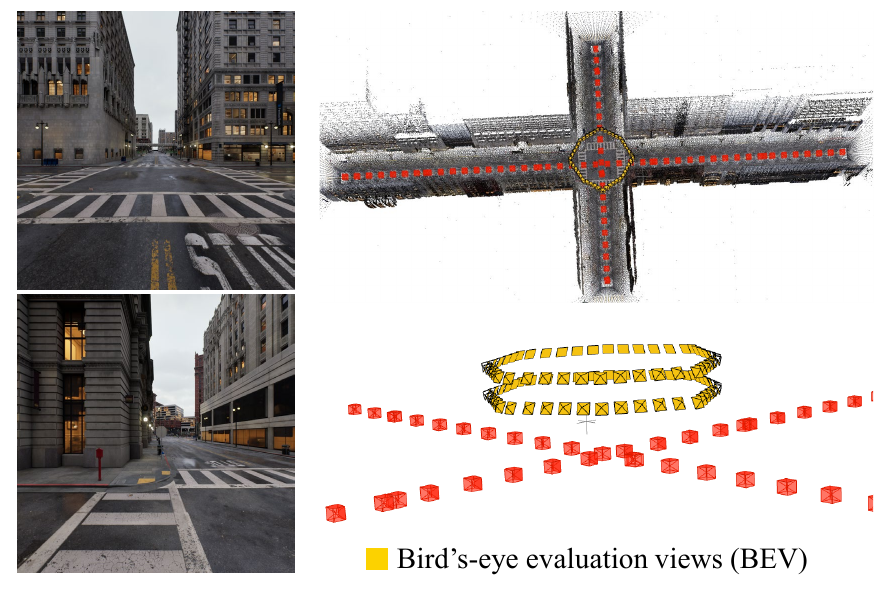}\\ \vspace{-2mm}
      \caption{Synthetic dataset of MatrixCity Block Small.}
\label{fig:synthetic_dataset}
\end{figure}

\noindent{\textbf{Dataset and settings:}} We used the synthetic Block Small dataset in MatrixCity~\cite{matrix_city}, as shown in Fig.~\ref{fig:synthetic_dataset}. The left images are examples of the input images. The top-right figure shows the point cloud and the camera poses. The Block Small dataset only provides street-level ($3m$ height and $0^\circ$ pitch) training and evaluation views shown as the red cameras, which are insufficient to evaluate the rendering quality from novel views far from the training views. Thus, we constructed additional evaluation views, shown as the yellow cameras in the bottom-right figure, for which the cameras were circularly moved at two different heights and pitches, i.e., ($10m$,~$25^\circ$) and ($13m$,~$30^\circ$). We call these sets of evaluation views bird's-eye evaluation views 1 and 2 (BEV\_1 and BEV\_2), respectively. In our method, we set $\Delta h = 10m$ and $\theta_{pitch} = 30^\circ$ for generating additional views, which are not identical to BEV\_1 and BEV\_2, because the additional views are generated according to the training views, as illustrated in Fig.~\ref{fig:added_pose}. We applied our proposed method (w/o and w/ diffusion) to three target 3DGS methods, i.e., Splatfacto~\cite{gsplat}, Scaffold-GS~\cite{scaffoldgs}, and 2DGS~\cite{2dgs}.

\begin{table}[!t]
    \caption{Quantitative comparisons on Block Small dataset. The bold represents the best, and the underline represents the second best for each target 3DGS method.}
    \vspace{-2mm}
    \centering
    \fontsize{7.4pt}{10pt}\selectfont
    \setlength{\tabcolsep}{3pt}
    \begin{tabularx}{\columnwidth}{>{\raggedright\arraybackslash}X|ccc|ccc}
        \hline
         & \multicolumn{3}{c|}{PSNR $\uparrow$}
         & \multicolumn{3}{c}{SSIM $\uparrow$} \\
         & Street & BEV\textunderscore$1$ & BEV\textunderscore$2$
         & Street & BEV\textunderscore$1$ & BEV\textunderscore$2$ \\
        \hline\hline
        StreetNeRF~\cite{streetnerf} & 23.56 & 22.02 & 21.43 & 0.759 & 0.722 & 0.701 \\ \hline
        Splatfacto~\cite{gsplat} & 22.47 & 18.71 & 16.33 & \textit{\underline{0.782}} & 0.658 & 0.608  \\
        Ours w/o diffusion (Splat.) & \textbf{23.49} & \textit{\underline{22.24}} & \textit{\underline{21.75}} & \textbf{0.787} & \textbf{0.748} & \textbf{0.729} \\
        Ours w/ diffusion (Splat.)  & \textit{\underline{22.93}} & \textbf{22.56} & \textbf{21.87} & 0.781 & \textit{\underline{0.746}} & \textit{\underline{0.725}} \\
        \hline
        Scaffold-GS~\cite{scaffoldgs} & \textbf{24.39} & 21.70 & 20.99 & \textbf{0.839} & 0.739 & 0.696 \\
        Ours w/o diffusion (Scaff.) & 24.14 & \textit{\underline{22.18}} & \textit{\underline{21.51}} & \textit{\underline{0.826}} & \textit{\underline{0.762}} & \textit{\underline{0.735}} \\
        Ours w/ diffusion (Scaff.) & \textit{\underline{24.19}} & \textbf{22.92} & \textbf{22.38} & 0.817 & \textbf{0.764} & \textbf{0.742} \\ \hline
        2DGS~\cite{2dgs} & 22.80 & 19.98 & 17.75 & \textbf{0.788} & 0.699 & 0.665 \\
        Ours w/o diffusion (2DGS) & \textbf{23.14} & \textit{\underline{22.36}} & \textit{\underline{21.77}} & \textit{\underline{0.771}} & \textit{\underline{0.753}} & \textit{\underline{0.735}} \\
        Ours w/ diffusion (2DGS) & \textit{\underline{22.91}} & \textbf{22.64} & \textbf{22.09} & 0.767 & \textbf{0.755} & \textbf{0.739} \\
        \hline
    \end{tabularx}
    \label{tab:quantitative}
\end{table}

\begin{figure*}[t!]
      \centering
\includegraphics[width=0.96\linewidth]{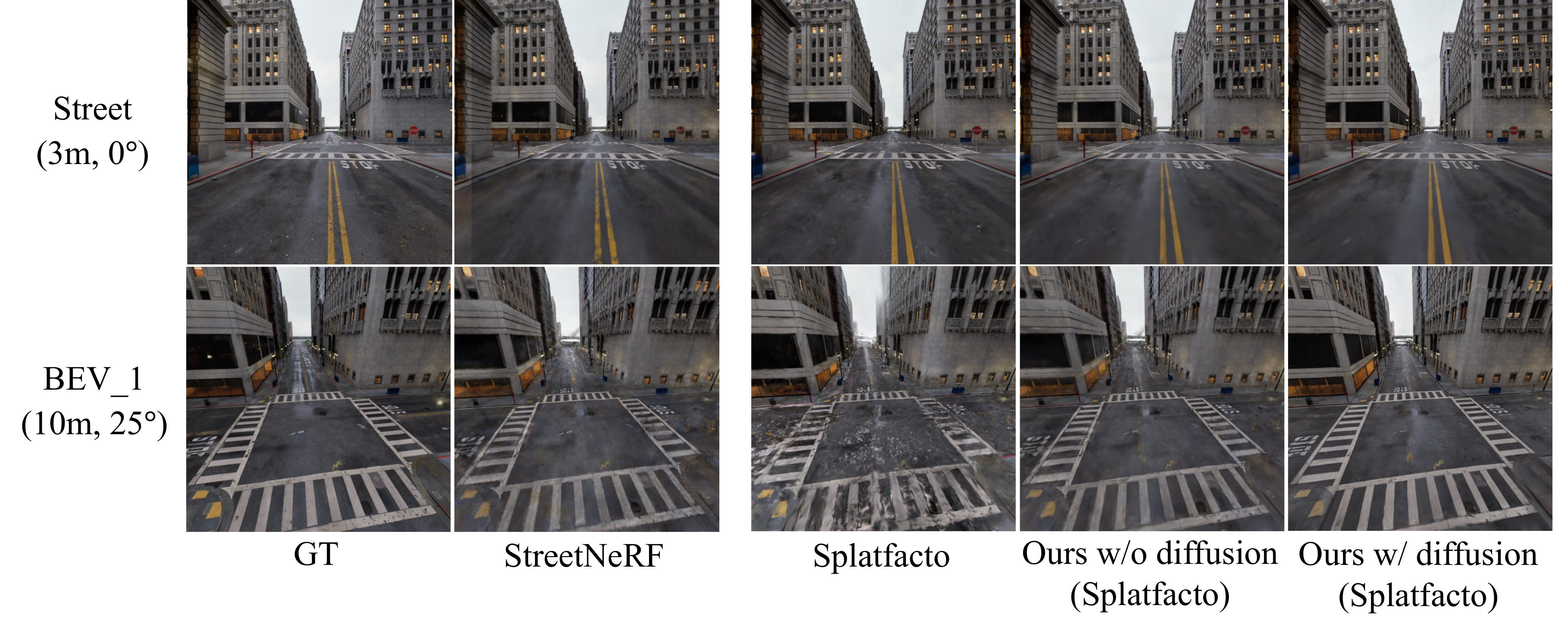}\\ \vspace{-4mm}
      \caption{Visual comparisons on the Block Small dataset using Splatfacto.}
\label{fig:visual}
\end{figure*}

\noindent{\textbf{Quantitative comparisons:}}  Table~\ref{tab:quantitative} shows the quantitative results, where we evaluate peak signal-to-noise ratio (PSNR) and structural similarity index measure (SSIM)~\cite{wang2004image}.  For the original street-level evaluation views (Street), our method does not consistently improve the original methods. This is because these views are close to the training views, resulting in sufficient quality even by the original methods. This is evident from that the numerical performance on BEV\_1 and BEV\_2 significantly drops for the original methods. In contrast, our methods w/o and w/ diffusion consistently improve the original methods on BEV\_1 and BEV\_2 and alleviate the performance drops compared with Street. Regarding the rendering speed, our method with Splatfacto maintains its original speed and achieves a frame-per-second of $50.86$ using an NVIDIA RTX 4090 GPU, which is much faster than that of StreetNeRF ($0.115$). These results indicate that our method realizes both high quality and high speed by combining NeRF and 3DGS.

\noindent{\textbf{Visual comparisons:}} Figure~\ref{fig:visual} shows the visual comparisons using Splatfacto. The visual results support a consistent interpretation with the numerical results. For Street on the top, all methods produce similar results sufficiently close to the ground truth~(GT). In contrast, for BEV\_1, Splatfacto generates significant artifacts and black-colored holes on the road. It is confirmed that the reference StreetNeRF succeeds in reducing those artifacts and holes, owing to the plane regularization of the road. Our method w/o diffusion produces a similar result with StreetNeRF, validating that our method can successfully inherit the rendering quality of StreetNeRF to the target Splatfacto. Finally, our method w/ diffusion demonstrates the best visual result, owing to the strong capability of the diffusion model to enhance image quality. The visual comparisons using Scaffold-GS and 2DGS are included in the supplementary material, where a similar tendency with Splatfacto can be observed.

\begin{table}[!t]
    \vspace{-3mm}
    \caption{Ablation study on Splatfacto.}
    \vspace{-2mm}
    \centering
    \setlength{\tabcolsep}{3pt}
    \small
    \begin{tabularx}{\columnwidth}{%
        c c c || >{\centering\arraybackslash}X >{\centering\arraybackslash}X >{\centering\arraybackslash}X}
        \hline
        \multicolumn{3}{c||}{} & \multicolumn{3}{c}{PSNR$\uparrow$} \\
        \makecell[c]{Add. views\\from 3DGS} &
        \makecell[c]{Add. views\\from NeRF} &
        Diffusion &
        Street & BEV\textunderscore$1$ & BEV\textunderscore$2$ \\ \hline
         & & & 22.47 & 18.71 & 16.33 \\
        \text{\checkmark} & & & \textit{\underline{22.97}} & 18.87 & 16.30 \\
        \text{\checkmark} & & \text{\checkmark} & 22.73 & 20.27 & 18.22 \\
         & \text{\checkmark} & & \textbf{23.49} & \textit{\underline{22.24}} & \textit{\underline{21.75}} \\
         & \text{\checkmark} & \text{\checkmark} & 22.93 & \textbf{22.56} & \textbf{21.87} \\ \hline
    \end{tabularx}
    \label{tab:ablation}
\end{table}

\begin{figure}
      \centering
      \vspace{-3mm}
\includegraphics[width=1.00\linewidth]{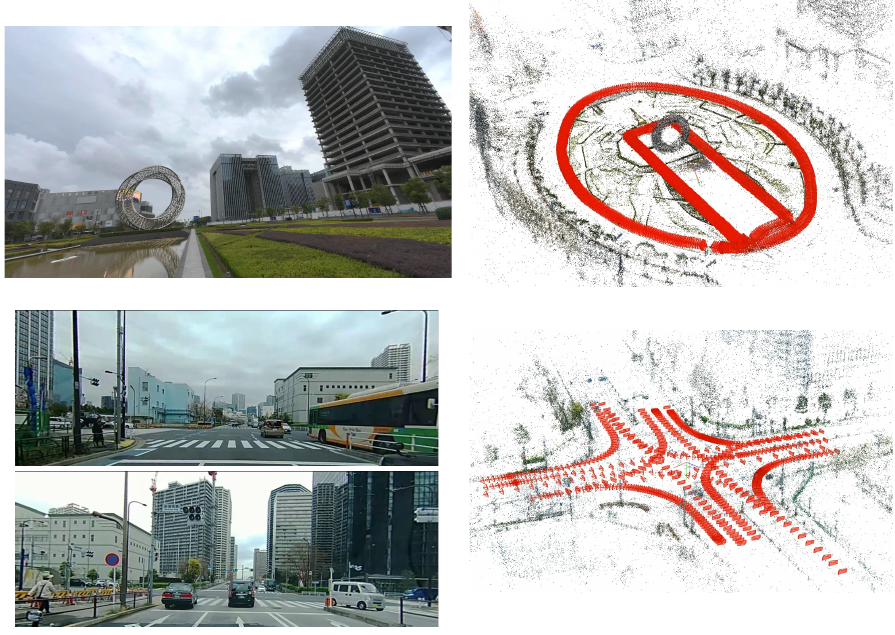}
      \caption{Real datasets of Park~(top) and Shinagawa~(bottom).}
\label{fig:real_dataset}
\end{figure}

\begin{figure*}[!t]
      \centering
\includegraphics[width=0.96\linewidth]{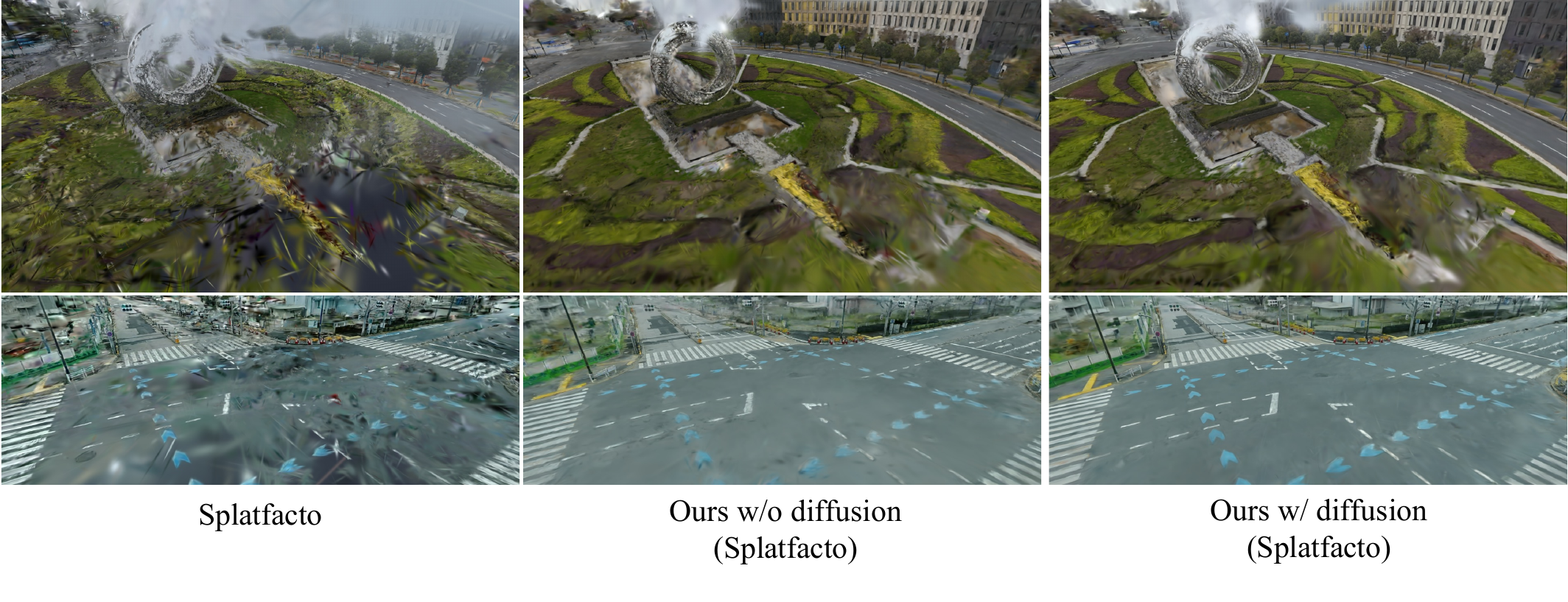}\\ \vspace{-5mm}
      \caption{Visual comparisons on the Park~(top) and the Shinagawa~(bottom) datasets using Splatfacto.}
\label{fig:qualitative}
\end{figure*}

\noindent{\textbf{Ablation study:}} To investigate the effectiveness of combining NeRF-rendered images and diffusion-based image enhancement, we conducted an ablation study shown in Table~\ref{tab:ablation} using Splatfacto. The first row shows the result of the original Splatfacto. The second row represents the method that utilizes the same additional views as ours, but they are rendered from the original Splatfacto. The third row is the method that further adopts the diffusion-based image enhancement to the additional views rendered from the Splatfacto. Compared with the first row, simply adding views rendered by the original Splatfacto (second row) yields no significant difference. While applying the diffusion-based image enhancement to these 3DGS-rendered views (third row) brings an improvement, our proposed methods leveraging NeRF-rendered additional views (the fourth and fifth rows) achieve a substantially larger performance improvement for BEV\_1 and BEV\_2. These results validate the importance of utilizing the NeRF method as the reference to generate the additional views.

\subsection{Results on real datasets}

\noindent{\textbf{Datasets and settings:}} We used two real datasets, as shown in Fig.~\ref{fig:real_dataset}. The first dataset is the Park dataset provided in~\cite{horizonGS}. While the Park dataset contains street-level and aerial images, we only used the street-level images for the training, according to our assumed situation in this study. Using only street-level images, we performed SfM by using COLMAP~\cite{schoenberger2016sfm} to estimate the camera poses and the point cloud. The parameters to generate additional views were empirically set as $\Delta h~=~5 \mathrm{m}$ and $\theta_{pitch} = 25^\circ$.

Another dataset shown in the bottom part of Fig.~\ref{fig:real_dataset} is our constructed dataset, named the Shinagawa dataset, capturing an intersection of Shinagawa City using a dashcam. The dataset includes a total of 12 video clips, i.e., four straight directions (South to North, North to South, West to East, and East to West) and eight turning directions (e.g., South to West, South to East, etc.). The camera poses and the point cloud were estimated by SfM using COLMAP. The parameters to generate additional views were empirically set as $\Delta h = 5 \mathrm{m}$ and $\theta_{pitch} = 15^\circ$.

\noindent{\textbf{Visual comparisons:}} Because there are no suitable evaluation views for the real datasets, we only performed visual comparisons for those datasets. Specifically, the aerial views in the original Park dataset are too high in height and placed outside the range of reconstruction areas from street-level views, making them unusable for evaluation views. Also, there are no views other than street-level views for the Shinagawa dataset because the images were captured using a dashcam mounted on a real car. 

Figure~\ref{fig:qualitative} presents visual comparisons on the Park~(top) and the Shinagawa~(bottom) datasets using Splatfacto. For both datasets, our proposed methods w/o and w/ diffusion successfully improve the original Splatfacto, significantly reducing the holes on the ground and the road, which is attributed to the incorporation of the plane regularization in StreetNeRF. Even though the Shinagawa dataset contains a large number of transient objects, artifacts caused by such objects are not apparent in the results of our methods, indicating that high-quality street-scene representations are successfully achieved, which is attributed to the exclusion of transient objects in the input views.

\subsection{Video results}

Video results can be confirmed from our project page\footnote{\url{http://www.ok.sc.e.titech.ac.jp/res/NVS/NeRFtoGS.html}}, including additional results using Scaffold-GS for real datasets. The overall trends of the results are consistent with the results in this paper, clearly highlighting the better rendering quality of our proposed methods in both cases w/o and w/ diffusion. 


\section{Conclusion}

In this paper, we have proposed a method for integrating a reference NeRF method to improve a target 3DGS method. As the reference NeRF method, we adopted StreetNeRF, which incorporates street-specific processing to achieve high-quality NVS. We used the trained StreetNeRF model to generate both input-view images and additional-view images. Image composition and diffusion-based image enhancement are applied to these images, respectively, and the resulting images are then used to train the target 3DGS method.
Both quantitative and qualitative evaluations have demonstrated that our proposed method achieves NVS quality comparable to or surpassing that of StreetNeRF, while maintaining the rendering speed of the original 3DGS method. One of the limitations of our method is that we currently generate the poses of additional views in an empirical manner. The generation of optimal poses for additional views is one of our future works.

\vspace{1em}
\noindent \textbf{Acknowledgment} This research was conducted as part of a collaborative research project with MICWARE CO., LTD.

\bibliographystyle{IEEEbib}
\bibliography{example}

\clearpage 

\twocolumn[
    \begin{center}
        \vspace*{5mm} 
        \Large \textbf{Supplementary Material: \\ Leveraging NeRF-Rendered Images for 3D Gaussian Splatting}
        \vspace{8mm} 
    \end{center}
]

\setcounter{figure}{0}
\setcounter{table}{0}
\renewcommand{\thefigure}{S\arabic{figure}}
\renewcommand{\thetable}{S\arabic{table}}

\section{Additional Results}

\begin{figure*}[!h]
    \centering
    \includegraphics[width=0.95\linewidth]{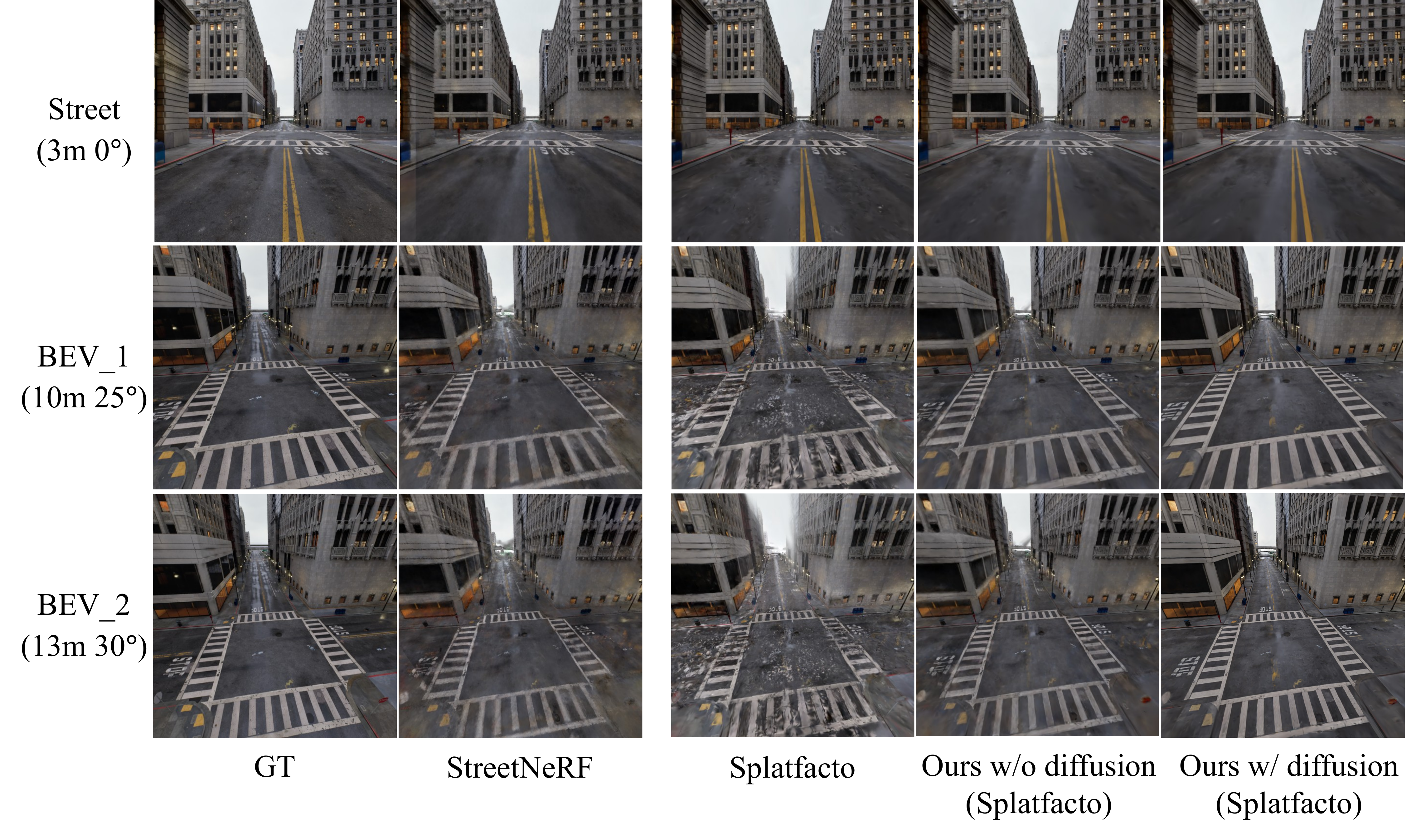}\\ \vspace{-4mm}
    \caption{Visual comparisons on the Block Small dataset using Splatfacto}
    \label{fig:sup_quantitative_splatfacto}
\end{figure*}

\begin{figure*}[!h]
    \centering
    \includegraphics[width=0.95\linewidth]{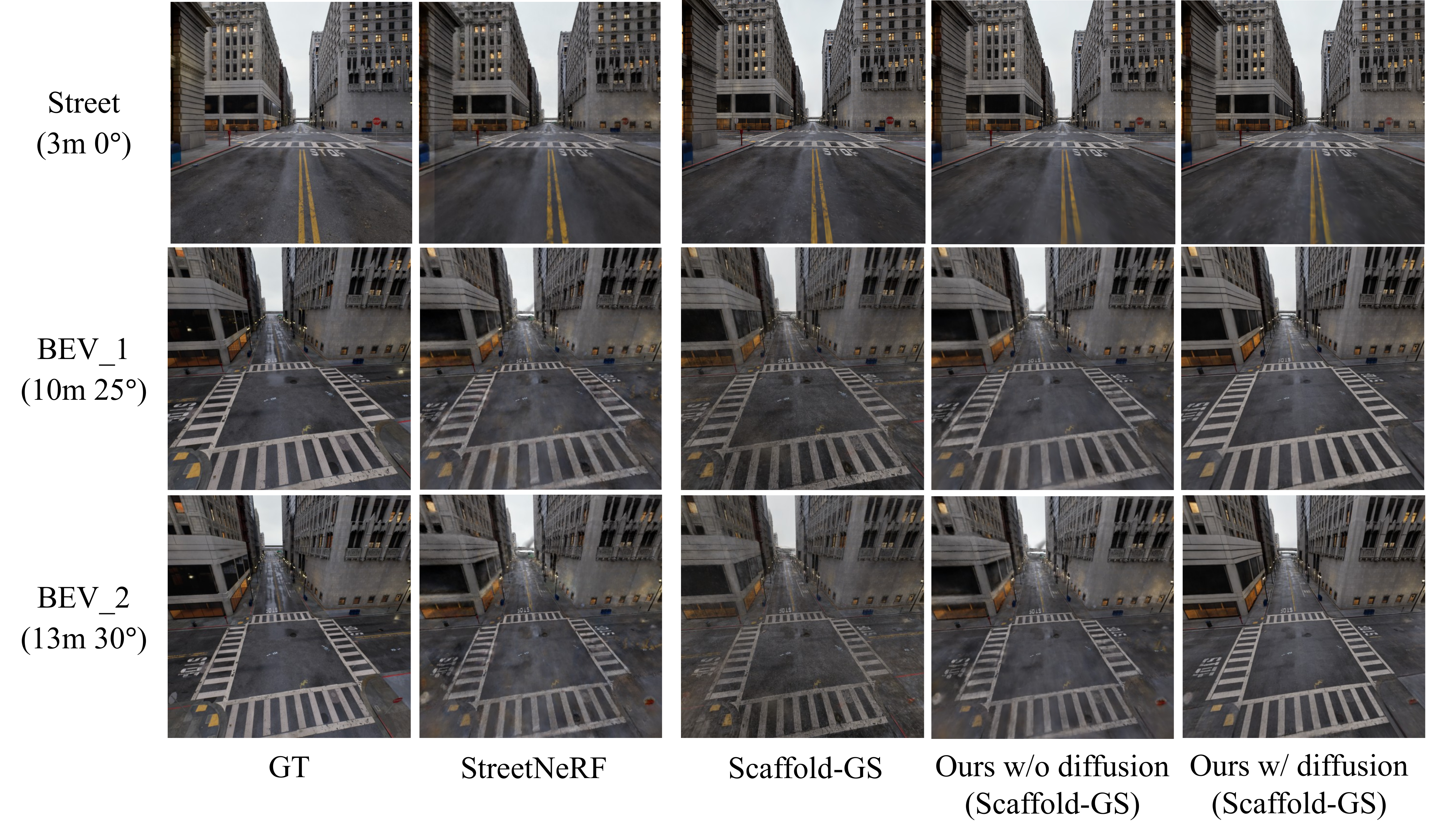}\\ \vspace{-4mm}
    \caption{Visual comparisons on the Block Small dataset using Scaffold-GS}
    \label{fig:sup_quantitative_scaffold}
\end{figure*}

\begin{figure*}[!h]
    \centering
    \includegraphics[width=0.95\linewidth]{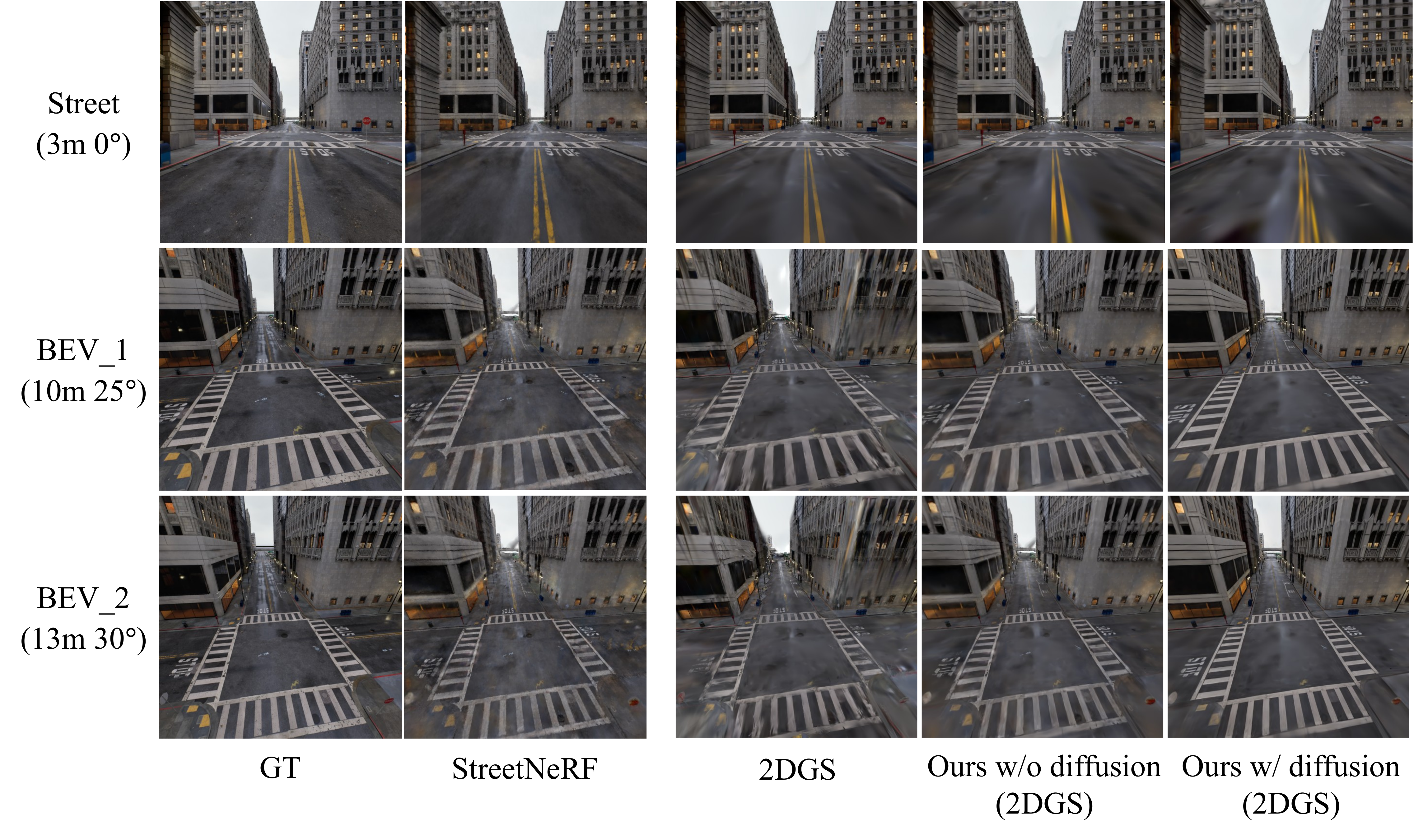}\\ \vspace{-4mm}
    \caption{Visual comparisons on the Block Small dataset using 2DGS}
    \label{fig:sup_quantitative_2dgs}
\end{figure*}

Table~\ref{tab:quantitative_supp} shows the full numerical comparison for the synthetic Block Small dataset, including the learned perceptual image patch similarity~(LPIPS)~\cite{zhang2018unreasonable}. The results of LPIPS for BEV\_1 and BEV\_2 indicate that our proposed method with diffusion enhancement shows the best performance for all 3DGS methods. 
Figures~\ref{fig:sup_quantitative_splatfacto} to~\ref{fig:sup_quantitative_2dgs} show additional visual comparisons for the Block Small dataset, including the results using Scaffold-GS and 2DGS, as well as the results for BEV\_2, which cannot be included in the main manuscript due to limited space. The overall trends of the results are consistent with those mentioned in the main manuscript. 

\begin{table*}[!h]
    \caption{Quantitative comparisons on Block Small dataset. The bold represents the best, and the underline represents the second best for each target 3DGS method.}
    \vspace{-2mm}
    \centering
    \setlength{\tabcolsep}{8pt} 
    \begin{tabular}{l|ccc|ccc|ccc} 
        \hline
         & \multicolumn{3}{c|}{PSNR $\uparrow$}
         & \multicolumn{3}{c|}{SSIM $\uparrow$}
         & \multicolumn{3}{c}{LPIPS $\downarrow$} \\
         & Street & BEV\textunderscore$1$ & BEV\textunderscore$2$
         & Street & BEV\textunderscore$1$ & BEV\textunderscore$2$
         & Street & BEV\textunderscore$1$ & BEV\textunderscore$2$ \\
        \hline\hline
        StreetNeRF~\cite{streetnerf} & 23.56 & 22.02 & 21.43 & 0.759 & 0.722 & 0.701 & 0.371 & 0.298 & 0.318 \\ \hline
        Splatfacto~\cite{gsplat} & 22.47 & 18.71 & 16.33 & \textit{\underline{0.782}} & 0.658 & 0.608 & \textbf{0.316} & 0.332 & 0.364 \\
        Ours w/o diffusion (Splat.) & \textbf{23.49} & \textit{\underline{22.24}} & \textit{\underline{21.75}} & \textbf{0.787} & \textbf{0.748} & \textbf{0.729} & \textit{\underline{0.365}} & \textit{\underline{0.299}} & \textit{\underline{0.312}} \\
        Ours w/ diffusion (Splat.)  & \textit{\underline{22.93}} & \textbf{22.56} & \textbf{21.87} & 0.781 & \textit{\underline{0.746}} & \textit{\underline{0.725}} & 0.369 & \textbf{0.263} & \textbf{0.269} \\ \hline
        Scaffold-GS~\cite{scaffoldgs} & \textbf{24.39} & 21.70 & 20.99 & \textbf{0.839} & 0.739 & 0.696 & \textbf{0.194} & \textit{\underline{0.215}} & \textit{\underline{0.249}} \\
        Ours w/o diffusion (Scaff.) & 24.14 & \textit{\underline{22.18}} & \textit{\underline{21.51}} & \textit{\underline{0.826}} & \textit{\underline{0.762}} & \textit{\underline{0.735}} & \textit{\underline{0.261}} & 0.264 & 0.290 \\
        Ours w/ diffusion (Scaff.) & \textit{\underline{24.19}} & \textbf{22.92} & \textbf{22.38} & 0.817 & \textbf{0.764} & \textbf{0.742} & 0.267 & \textbf{0.208} & \textbf{0.216} \\ \hline
        2DGS~\cite{2dgs} & 22.80 & 19.98 & 17.75 & \textbf{0.788} & 0.699 & 0.665 & \textbf{0.382} & 0.350 & 0.371 \\
        Ours w/o diffusion (2DGS) & \textbf{23.14} & \textit{\underline{22.36}} & \textit{\underline{21.77}} & \textit{\underline{0.771}} & \textit{\underline{0.753}} & \textit{\underline{0.735}} & \textit{\underline{0.426}} & \textit{\underline{0.323}} & \textit{\underline{0.334}} \\
        Ours w/ diffusion (2DGS) & \textit{\underline{22.91}} & \textbf{22.64} & \textbf{22.09} & 0.767 & \textbf{0.755} & \textbf{0.739} & 0.427 & \textbf{0.301} & \textbf{0.305} \\ \hline
    \end{tabular}
    \label{tab:quantitative_supp}
\end{table*}


\end{document}